\documentclass[11pt,a4paper]{article}

\usepackage[margin=1in]{geometry}
\usepackage{amsmath,amssymb}
\usepackage{graphicx}
\usepackage{booktabs}
\usepackage{xcolor}
\usepackage[colorlinks=true,allcolors=blue]{hyperref}
\usepackage{url}
\usepackage{cite}
\usepackage{authblk}
\usepackage{subcaption}
\usepackage{comment}

\title{Eye laterality classification from retinal laser Doppler holography images using YOLO}
\author{Matthis Guillet, Paul Witkowski, Raphaël Sauvage-Marchin, Marius Dubosc, Michael Atlan}
\date{\today}

\begin{document}
\maketitle

\begin{abstract}
Quantitative retinal hemodynamic analysis requires correct anatomical orientation for longitudinal consistency and clinical reporting. Traditionally, laterality (left vs.\ right eye) has been recorded manually by operators, a process prone to human error that can compromise large-scale data analysis. We present an automated approach for laterality classification using an Ultralytics \textit{YOLO11n-cls} convolutional neural network trained on moment-0 (M0) structural images.
Our model correctly classified 82 of 84 images (97.62\%) in a held-out test set of 42 subjects, supporting automated metadata verification in Doppler holography pipelines. The proposed method enables automatic assignment of eye laterality metadata, reducing reliance on manual entry while providing an independent verification mechanism to detect potential labeling errors.
\end{abstract}

\section{Introduction}
Laser Doppler holography (LDH) enables retinal blood flow imaging and the estimation of arterial and venous blood velocity waveforms \cite{Pellizzari2016,Puyo2018,Puyo2019Waveforms,Martinache2026,EyeFlow,Auray2025Quantitative,Auray2026VelocityProxy}. A critical prerequisite for interpreting these hemodynamic biomarkers is the accurate assignment of eye laterality. The relative positions of the optic disc and fovea—and consequently the orientation of the major vascular arcades—are mirrored between the left (OS) and right (OD) eyes.

In clinical practice, laterality is often entered manually into acquisition software. However, in high-throughput research environments or multi-center trials, manual metadata entry is a persistent source of error \cite{zapata2020opth}. Mislabeled laterality can lead to incorrect hemispheric comparisons, failures in longitudinal image registration, and invalidation of study cohorts.

Automated laterality detection has been extensively explored in standard color fundus photography \cite{kang2022asymmetry, jang2018laterality, liu2018largescale} and other modalities such as anterior segment photography \cite{zheng2021anterior}. Historically, these methods relied on explicit landmark detection, inferring orientation from the geometric relationship between the optic disc and the fovea \cite{niemeijer2009odfovea}. However, in LDH acquisitions, the field of view (FOV) is typically restricted to the optic disc and its immediate surroundings to maximize sampling rates for hemodynamic analysis. Consequently, the fovea is rarely visible in the structural M0 images, rendering dual-landmark geometric rules inapplicable.

To address this challenge, we use structural asymmetries, such as vascular arcade curvature, in LDH moment-zero (M0) images generated with \textit{HoloDoppler} (versions 0.7 and 1.0) \cite{HoloDoppler}. We evaluate YOLO11 classification architectures \cite{ultralytics_yolo11_docs} and integrate the selected model into \textit{DopplerView} (version 1.8.1) \cite{DopplerView} for automated laterality assignment and verification.

\section{Materials and Methods}

\subsection{Holographic Image Acquisition and Reconstruction}

Retinal interferograms were acquired using a standardized laser Doppler holography instrument based on an inline Mach--Zehnder interferometer operating at a wavelength of 852\,nm. Diffuse Maxwellian illumination was used, and cross-polarized light backscattered from the fundus was combined with a reference beam. Interferograms were recorded at 37,000 frames per second using a high-speed CMOS camera \cite{Puyo2018,Auray2025Quantitative,bratasz_diffuse_2024}.

\begin{figure}[htbp]
    \centering
    \captionsetup[sub]{font=footnotesize}
    \begin{subfigure}[b]{0.24\textwidth}
        \centering
        \includegraphics[width=\linewidth]{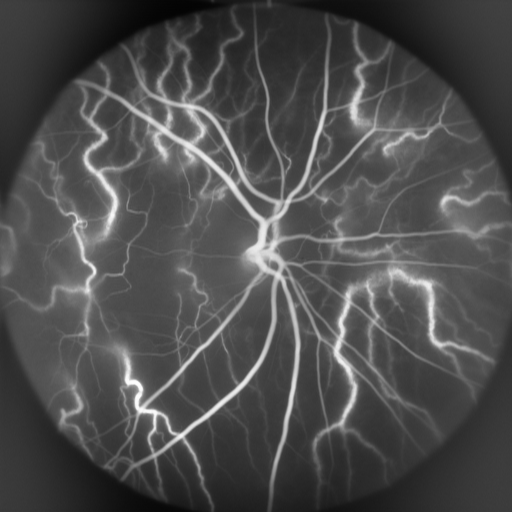}
        \subcaption{Right eye}
    \end{subfigure}
    \hfill
    \begin{subfigure}[b]{0.24\textwidth}
        \centering
        \includegraphics[width=\linewidth]{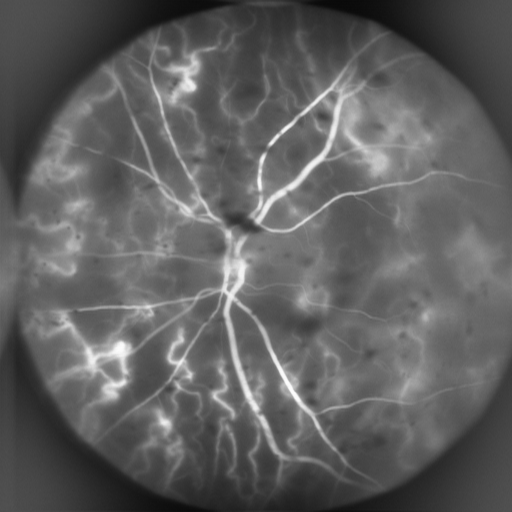}
        \subcaption{Left eye}
    \end{subfigure}
    \hfill
    \begin{subfigure}[b]{0.24\textwidth}
        \centering
        \includegraphics[width=\linewidth]{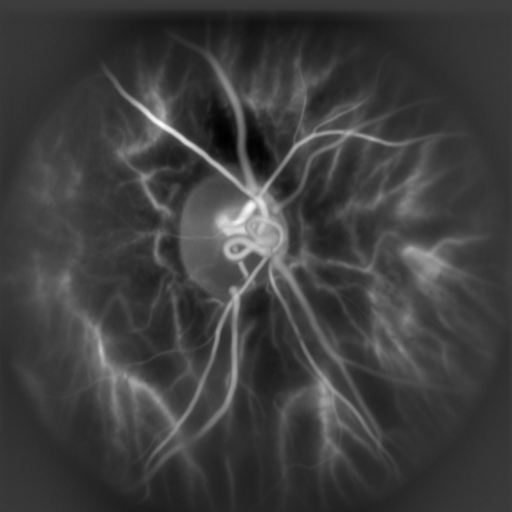}
        \subcaption{Right eye}
    \end{subfigure}
    \hfill
    \begin{subfigure}[b]{0.24\textwidth}
        \centering
        \includegraphics[width=\linewidth]{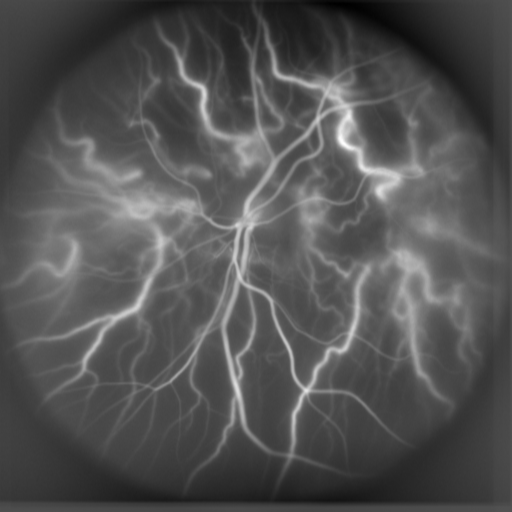}
        \subcaption{Left eye}
    \end{subfigure}
    
    \vspace{0.2cm} 
    
    \begin{subfigure}[b]{0.24\textwidth}
        \centering
        \includegraphics[width=\linewidth]{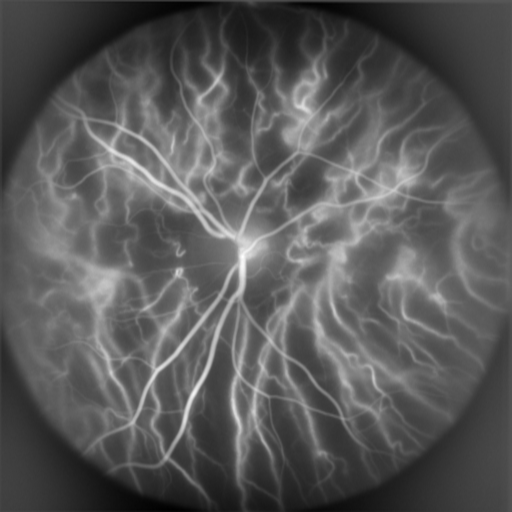}
        \subcaption{Right eye}
    \end{subfigure}
    \hfill
    \begin{subfigure}[b]{0.24\textwidth}
        \centering
        \includegraphics[width=\linewidth]{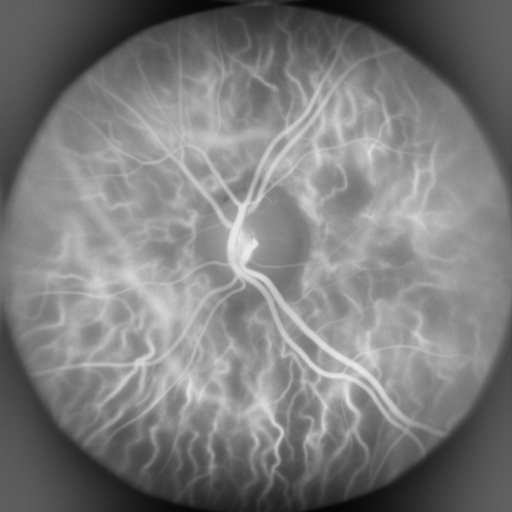}
        \subcaption{Left eye}
    \end{subfigure}
    \hfill
    \begin{subfigure}[b]{0.24\textwidth}
        \centering
        \includegraphics[width=\linewidth]{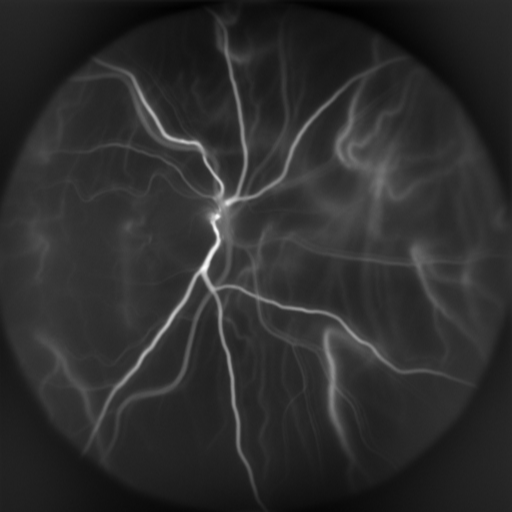}
        \subcaption{Right eye}
    \end{subfigure}
    \hfill
    \begin{subfigure}[b]{0.24\textwidth}
        \centering
        \includegraphics[width=\linewidth]{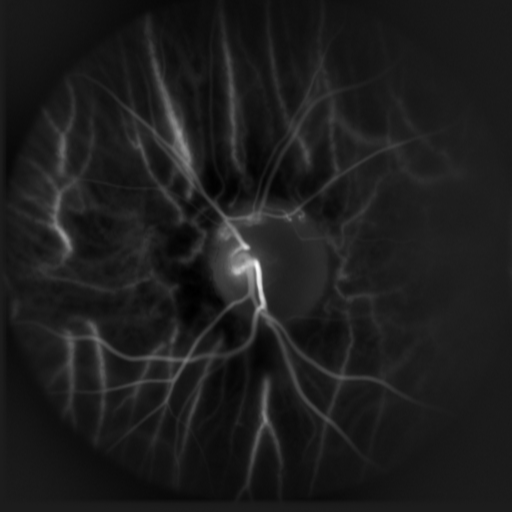}
        \subcaption{Left eye}
    \end{subfigure}
    
    \caption{Representative Doppler images used for laterality classification. The images illustrate variations in vascular geometry and optic disc appearance. Each image corresponds to a different subject; the eyes are not shown in matched pairs.}
    \label{fig:m0_grid}
\end{figure}

Raw interferograms were processed using the \textit{HoloDoppler} rendering pipeline \cite{HoloDoppler}. Fresnel propagation, singular value decomposition filtering, and short-time Fourier transformation were used to reconstruct spatially resolved Doppler power spectra \cite{puyo_spatio-temporal_2020}. Integration of the spectral power over a selected frequency band yielded moment-zero ($M_0$) images. These power Doppler images reveal the optic nerve head and retinal vascular architecture, providing anatomical cues for eye laterality classification (Fig.~\ref{fig:m0_grid}).

\subsection{Dataset and Subject-Level Partitioning}
\label{sec:dataset}

A clinical dataset was constructed from Doppler holography M0 (power Doppler) images, each consisting of a grayscale rendering of the retinal anatomy. All images were acquired at the Quinze-Vingts National Eye Hospital in Paris, France. Written informed consent was obtained from all subjects prior to inclusion. The classification task consists of identifying the laterality (left or right) of the imaged eye.

The dataset comprises 92 subjects aged 20 to 60 years, presenting a wide variety of ocular diseases. For each subject, a single visit was included, during which both the left and right eyes were imaged within a short time interval. Since an eye retains most of its anatomical characteristics over time, including multiple visits of the same eye would add little new information and could lead the model to memorize individual eyes rather than learn features that distinguish left from right. We therefore chose to maximize the number of distinct subjects rather than the number of images per eye, in order to favor generalization. In total, 184 M0 images were used for classification.

Eye laterality was initially recorded by the examining ophthalmologist at the time of acquisition, based on whether the patient's right (OD) or left (OS) eye was being imaged. During dataset construction, these labels were reviewed by an expert (Michael Atlan) in laser Doppler holography (LDH) before being retained as reference annotations for model training and evaluation.

The dataset was split at the subject level into a development set and a held-out test set, so that both eyes of a given subject always belong to the same set. The development set was used for five-fold cross-validation and final model training; the held-out test set was reserved for final evaluation. The distribution of subjects and images across the two sets is detailed in Table~\ref{tab:dataset}.

\begin{table}[h]
\centering
\caption{Dataset composition and subject-level partitioning.}
\label{tab:dataset}
\begin{tabular}{lccc}
\toprule
Partition & Subjects & Images ($N$) & Purpose \\
\midrule
Development & 50 & 100 & $5$-fold cross-validation \\
Test     & 42 & 84  & Final evaluation \\
\midrule
Total    & 92 & 184 & \\
\bottomrule
\end{tabular}
\end{table}

During training, the Ultralytics pipeline applies extensive automated data augmentation (e.g., scaling), to promote generalization to unseen images~\cite{UltralyticsTrain}.

\subsection{Deep Learning Model and Training}
We evaluated the YOLO11 classification architectures from the Ultralytics framework \cite{ultralytics_yolo11_docs} on $512 \times 512$-pixel M0 images. While \textit{YOLO} is primarily known for object detection, its backbone enables efficient feature extraction for whole-image classification tasks \cite{ultralytics_classify_docs}.

Networks were initialized with weights pretrained on \textit{ImageNet} \cite{ultralytics_imagenet}. Training was performed on an NVIDIA RTX 5090 GPU for 100 epochs using cross-entropy loss, stochastic gradient descent (SGD), and a batch size of 16. No preprocessing beyond that of the YOLO11 classification pipeline was applied. Horizontal flip augmentation was disabled to preserve the anatomical relationship between image orientation and laterality labels \cite{kang2022asymmetry}. For reproducibility, the complete training configuration, including hyperparameters and data augmentation settings, is publicly available in our GitHub repository \cite{EyeSideClassification}.

\subsection{Cross-Validation and Learning-Curve Evaluation}

We performed five-fold subject-level cross-validation on the development dataset (50 subjects, 100 images). This procedure was used to compare YOLO11 architectures (n, s, m, l, and x) and select the model and hyperparameters. Each fold contained 10 subjects (20 images). At each iteration, four folds were used for training and one for validation, so no subject appeared in both sets. Models were compared by mean validation accuracy and macro F1-score. Ties were resolved in favor of the smallest and fastest model. Inference time was measured by running each model 100 times on the same image. The final model was then trained on the full development dataset with the selected settings.
For model evaluation, the final training checkpoint (last.pt) was used instead of the validation-selected checkpoint (best.pt), avoiding checkpoint selection based on performance on the evaluated validation fold.

To investigate the effect of training-set size, a learning-curve analysis was conducted using the same five folds. For each fold, nested training subsets of increasing size, ranging from 2 to 80 images (1 to 40 subjects), were sampled at the subject level, with smaller training sizes emphasized through an approximately logarithmic progression. The validation partition remained fixed across all training-set sizes within each fold, allowing direct comparisons under identical evaluation conditions.

The test dataset (42 subjects, 84 images) was not used for cross-validation, model selection, or the learning curve. It was reserved for final evaluation.
\section{Results}

\subsection{Cross-Validation Performance}

Table~\ref{tab:cv_comparison} compares the five YOLO11 classification architectures. YOLO11n-cls showed the best overall performance, combining the highest mean validation accuracy ($99.0 \pm 2.0\%$) and macro F1-score ($99.0 \pm 2.0\%$) with the fewest parameters (1.534~M) and the shortest inference time ($1.483 \pm 0.097$~ms). YOLO11s-cls and YOLO11l-cls reached the same accuracy and F1-score but required 3.6 and 8.4 times more parameters and were 1.2 and 3.0 times slower, respectively. YOLO11m-cls and YOLO11x-cls had a lower mean accuracy ($98.0 \pm 2.4\%$). YOLO11n-cls was therefore retained as the final architecture.

For YOLO11n-cls, 99 out of 100 out-of-fold predictions were correct, with fold accuracies ranging from 95.0\% to 100.0\%. Macro-averaged precision, recall, and F1-score reached 99.1\%, 99.0\%, and 99.0\%, respectively. Because cross-validation was also used for model selection, these values are development estimates; performance on unseen subjects is assessed separately on the held-out test set (Section~\ref{sec:test_performance}).

\begin{table}[htbp]
    \centering
    \caption{Comparison of YOLO11 classification architectures
    using five-fold subject-level cross-validation.
    Accuracy and macro F1-score are reported as
    mean $\pm$ standard deviation across folds.
    Parameter counts correspond to models adapted
    for binary eye laterality classification.
    Inference time is reported as mean $\pm$ standard deviation over 100 repeated inferences on a single image, using the same image for all models.}
    
    \label{tab:cv_comparison}

    \small
    \begin{tabular}{lcccc}
        \toprule
        \textbf{Model}
        & \textbf{Accuracy (\%)}
        & \textbf{F1 (\%)}
        & \textbf{Params (M)}
        & \textbf{Time (ms)} \\
        \midrule
        \textbf{YOLO11n-cls} & \textbf{99.0 ± 2.0} & \textbf{99.0 ± 2.0} & \textbf{1.534}  & \textbf{1.483 ± 0.097} \\
        YOLO11s-cls & 99.0 ± 2.0 & 99.0 ± 2.0 & 5.446  & 1.711 ± 0.060 \\
        YOLO11m-cls & 98.0 ± 2.4 & 98.0 ± 2.5 & 10.356 & 3.211 ± 0.010 \\
        YOLO11l-cls & 99.0 ± 2.0 & 99.0 ± 2.0 & 12.837 & 4.392 ± 0.012 \\
        YOLO11x-cls & 98.0 ± 2.4 & 98.0 ± 2.5 & 28.359 & 8.100 ± 0.012 \\
        \bottomrule
    \end{tabular}
\end{table}

\subsection{Learning Curve Analysis}

\begin{figure}[htbp]
    \centering
    \includegraphics[width=1\linewidth]{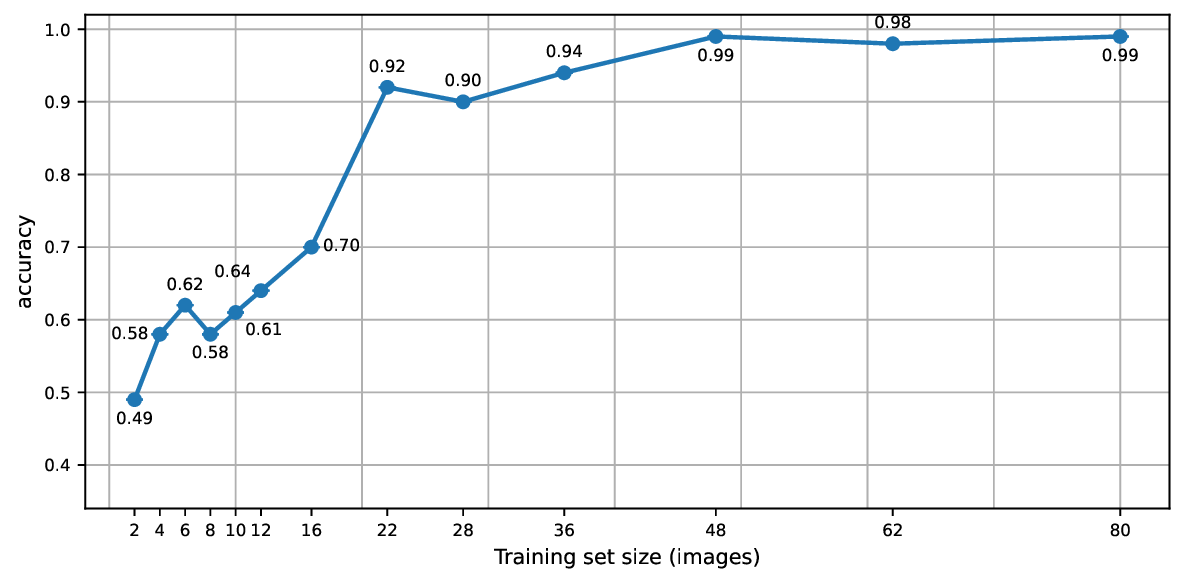}
    
    \caption{Learning curve of  \textit{YOLO11n-cls} obtained using five-fold subject-level cross-validation. The curve represents the mean classification accuracy across three repetitions, computed by aggregating out-of-fold predictions for each training-set size. The number of training images is displayed on a logarithmic scale. The same validation folds were maintained throughout the experiment to ensure consistent evaluation conditions.}
    \label{fig:learning_curve}
\end{figure}

To assess the influence of training-set size on classification performance, a learning curve was constructed using five-fold subject-level cross-validation on the development dataset (100 images from 50 subjects). For each fold, the model was independently trained on subject-level subsets of increasing size, ranging from 2 to 80 images, and evaluated on a fixed validation set of 20 images. Each training-set size was evaluated three times using different random subject selections, and the resulting accuracies were averaged across folds and repetitions.

As shown in Figure~\ref{fig:learning_curve}, classification accuracy increases rapidly with the number of training images, exceeding approximately 92\% with 22 images (11 subjects). Beyond this point, performance improves more gradually, reaching approximately 100\% with 80 training images (40 subjects). These results suggest that discriminative features for eye laterality can be learned from relatively small training datasets, with diminishing performance gains as additional training images are introduced.

\subsection{Classification Performance}
\label{sec:test_performance}
The final YOLO11n-cls model was evaluated on the independent test set of 84 images (42 subjects). It correctly classified 82 of 84 images (97.62\%). Detailed performance metrics are provided in Table~\ref{tab:metrics}.

\begin{table}[h]
\centering
\small
\caption{Image-level laterality classification metrics on the held-out test set.}
\label{tab:metrics}
\begin{tabular}{@{}lcccc@{}}
\toprule
Class & Precision & Recall & F1-Score & Support \\ \midrule
Right Eye & 0.9762 & 0.9762 & 0.9762 & 42 \\
Left Eye  & 0.9762 & 0.9762 & 0.9762 & 42 \\ \midrule
\textbf{Accuracy} & & & \textbf{0.9762} & \textbf{84} \\
\bottomrule
\end{tabular}
\end{table}

The confusion matrix (Table~\ref{tab:confusion}) shows 2 misclassifications out of 84 images: one right eye was predicted as left, and one left eye was predicted as right. Qualitative review identifies two possible contributing factors. The first image (Figure~\ref{fig:failure_2}) shows severe blur consistent with eye movement during acquisition. The M0 images are obtained by averaging power Doppler renderings over time. Saccadic or drift motion therefore creates ghosting and blurring, which degrade the global structure of the image. In the second image (Figure~\ref{fig:failure_1}), the vessels remain sharp, but the optic disc is not centered. It lies in the left part of the image, whereas it is usually close to the center. This unusual position may have altered the spatial pattern used by the model.

\begin{table}[h]
\centering
\small
\caption{Confusion matrix for the held-out test set.}
\label{tab:confusion}
\begin{tabular}{lcc}
\toprule
 & Predicted Right & Predicted Left \\ \midrule
\textbf{Actual Right} & 41 & 1 \\
\textbf{Actual Left} & 1 & 41 \\ \bottomrule
\end{tabular}
\end{table}

\begin{figure}[htbp]
    \centering
    \begin{subfigure}[b]{0.24\textwidth}
        \centering
        \includegraphics[width=\linewidth]{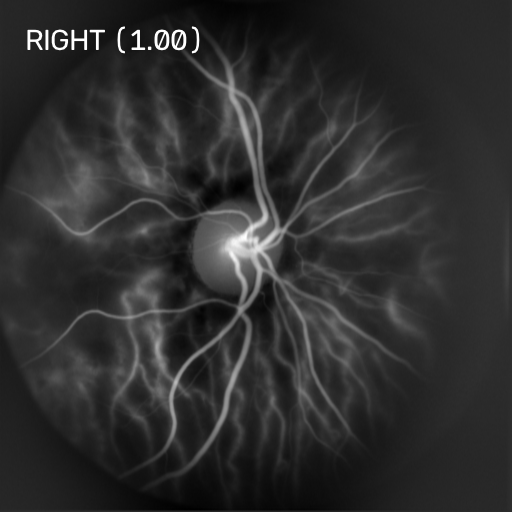}
        \subcaption{Right eye}
    \end{subfigure}
    \hfill
    \begin{subfigure}[b]{0.24\textwidth}
        \centering
        \includegraphics[width=\linewidth]{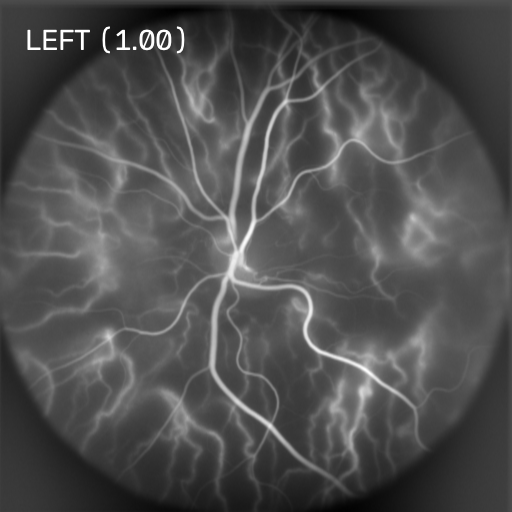}
        \subcaption{Left eye}
    \end{subfigure}
    \hfill
    \begin{subfigure}[b]{0.24\textwidth}
        \centering
        \includegraphics[width=\linewidth]{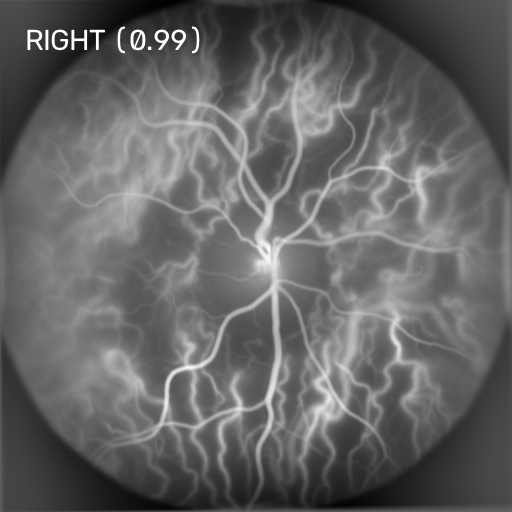}
        \subcaption{Right eye}
    \end{subfigure}
    \hfill
    \begin{subfigure}[b]{0.24\textwidth}
        \centering
        \includegraphics[width=\linewidth]{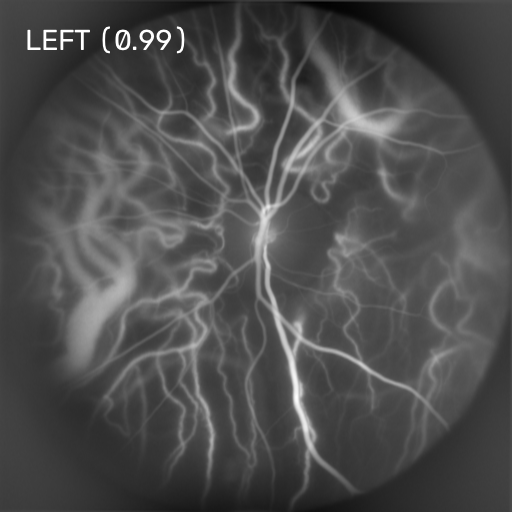}
        \subcaption{Left eye}
    \end{subfigure}
    
    \vspace{0.2cm} 
    
    \begin{subfigure}[b]{0.24\textwidth}
        \centering
        \includegraphics[width=\linewidth]{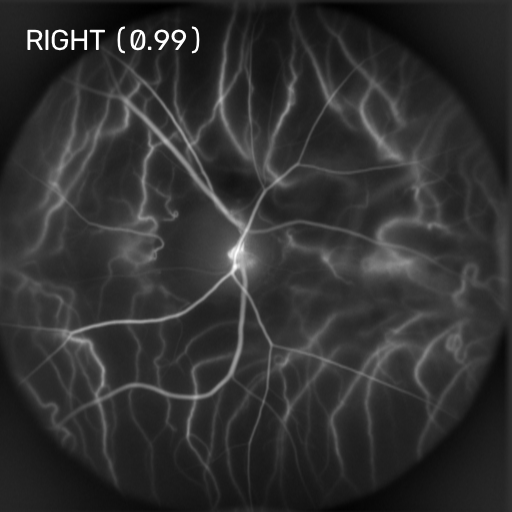}
        \subcaption{Right eye}
    \end{subfigure}
    \hfill
    \begin{subfigure}[b]{0.24\textwidth}
        \centering
        \includegraphics[width=\linewidth]{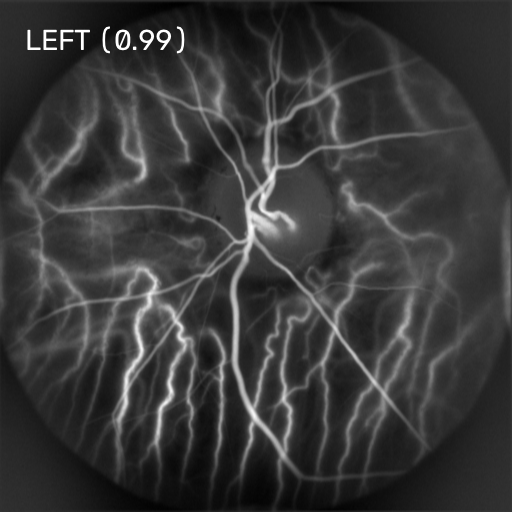}
        \subcaption{Left eye}
    \end{subfigure}
    \hfill
    \begin{subfigure}[b]{0.24\textwidth}
        \centering
        \includegraphics[width=\linewidth]{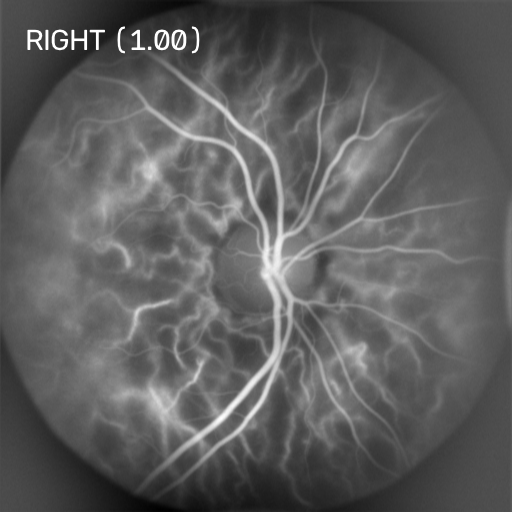}
        \subcaption{Right eye}
    \end{subfigure}
    \hfill
    \begin{subfigure}[b]{0.24\textwidth}
        \centering
        \includegraphics[width=\linewidth]{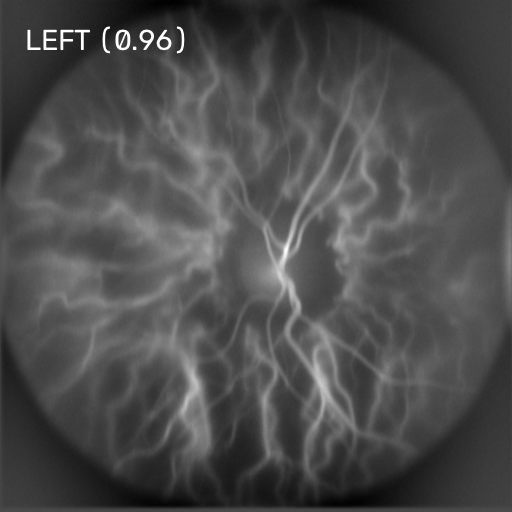}
        \subcaption{Left eye}
    \end{subfigure}
    
    \caption{Representative correctly classified test M0 images, with predicted laterality and model confidence overlaid.}
    \label{fig:m0_results}
\end{figure}

Figure~\ref{fig:m0_results} illustrates the practical use of the classifier. Test images were processed by the trained  \textit{YOLO11n-cls} model. 
These representative test images illustrate correct laterality predictions across variations in retinal appearance. Disease-specific performance was not assessed. Integrated into DopplerView, the classifier is intended to automatically verify laterality labels and reduce the risk of left/right mix-ups during image acquisition and archiving.

\subsection{Low-Confidence Predictions and Misclassifications}

\begin{figure}[!ht]
    \centering
    \captionsetup[subfigure]{font=footnotesize}

    \begin{subfigure}[t]{0.48\textwidth}
        \centering
        \begin{minipage}{0.49\linewidth}
            \centering
            \includegraphics[width=\linewidth,height=0.18\textheight,keepaspectratio]
                {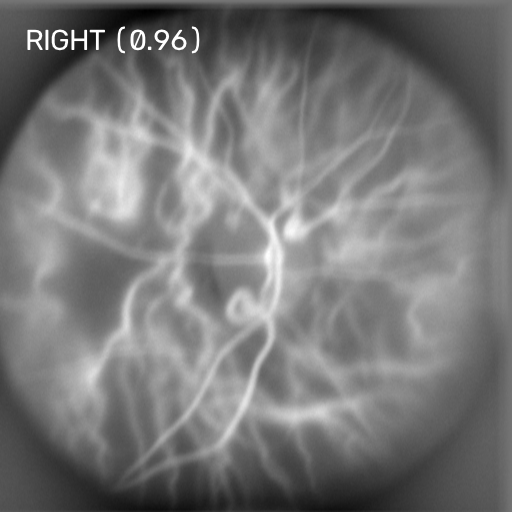}
        \end{minipage}\hfill
        \begin{minipage}{0.49\linewidth}
            \centering
            \includegraphics[width=\linewidth,height=0.18\textheight,keepaspectratio]
                {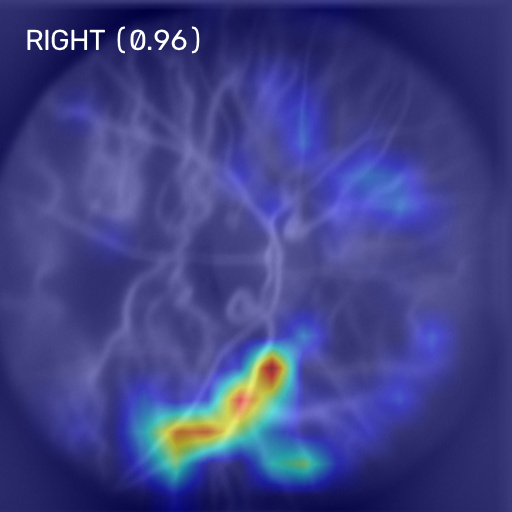}
        \end{minipage}
        \caption{Right eye; Low confidence (0.96)}
        \label{fig:failure_unsure1}
    \end{subfigure}
    \hfill
    \begin{subfigure}[t]{0.48\textwidth}
        \centering
        \begin{minipage}{0.49\linewidth}
            \centering
            \includegraphics[width=\linewidth,height=0.18\textheight,keepaspectratio]
                {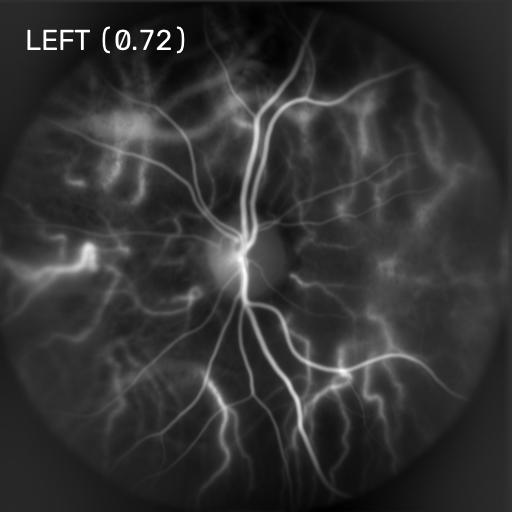}
        \end{minipage}\hfill
        \begin{minipage}{0.49\linewidth}
            \centering
            \includegraphics[width=\linewidth,height=0.18\textheight,keepaspectratio]
                {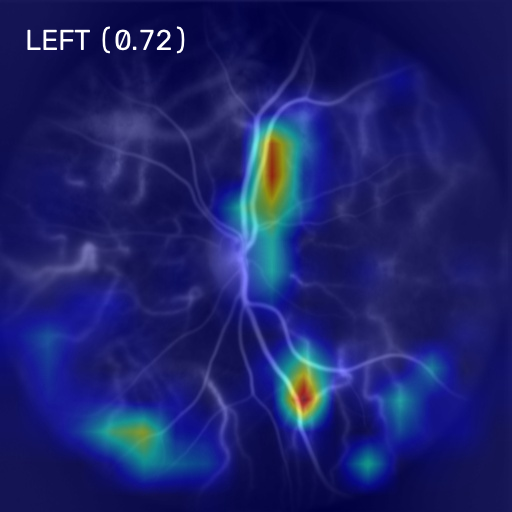}
        \end{minipage}
        \caption{Left eye; Low confidence (0.72)}
        \label{fig:failure_unsure2}
    \end{subfigure}

    \caption{The correctly classified test images with the lowest confidence scores. (a) A blurred, low-contrast image. (b) A sharp image with a lower confidence score, showing that blur alone does not explain the confidence score. Each M0 image is shown alongside its LayerCAM visualization, with the predicted laterality and confidence score. These maps may help explain the model’s predictions but do not establish their cause.}
    \label{fig:low_confidence}
    \label{fig:failure_analysis}
\end{figure}

Model confidence varied among correctly classified test images. Figure~\ref{fig:low_confidence} shows the two correctly classified images with the lowest confidence scores (0.96 and 0.72). Figure~\ref{fig:failure_unsure1} is blurred and has low contrast, so the vessels stand out less from the background. Figure~\ref{fig:failure_unsure2} is sharp, so blur does not explain its lower confidence. Image quality alone therefore does not explain model confidence.

Figure~\ref{fig:m0_failure} shows the two misclassified test images. Figure~\ref{fig:failure_2} is severely blurred, and the optic disc and retinal vessels are barely visible. The model still predicted the wrong laterality with high confidence (0.98). This example illustrates that a high confidence score can accompany an incorrect prediction on a degraded image. Figure~\ref{fig:failure_1} is sharp, but the optic disc lies in the left part of the image. Its confidence was lower (0.67), below that of all correctly classified test images.

In practice, blurred acquisitions are discarded and repeated because they do not provide reliable Doppler measurements. Laterality classification is performed only on images that pass quality control, which may reduce the risk of errors related to blur. This procedure does not address errors related to an off-center optic disc.

\subsection{Model Interpretability}

\begin{figure}[htbp]
    \centering
    \begin{subfigure}[b]{0.24\textwidth}
        \centering
        \includegraphics[width=\linewidth]{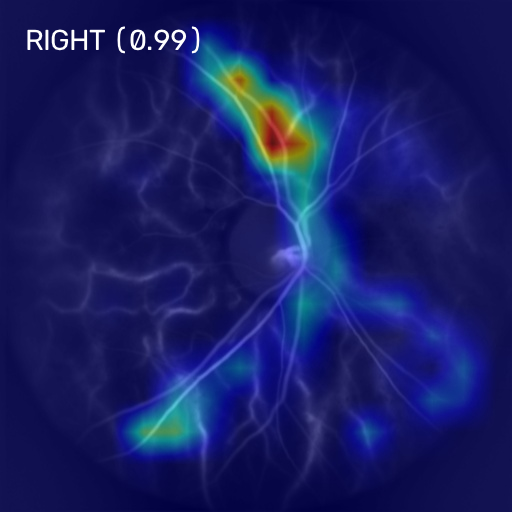}
        \subcaption{Right eye}
    \end{subfigure}
    \hfill
    \begin{subfigure}[b]{0.24\textwidth}
        \centering
        \includegraphics[width=\linewidth]{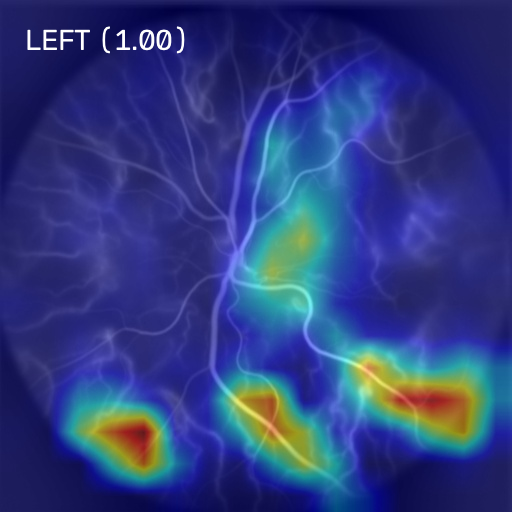}
        \subcaption{Left eye}
    \end{subfigure}
    \hfill
    \begin{subfigure}[b]{0.24\textwidth}
        \centering
        \includegraphics[width=\linewidth]{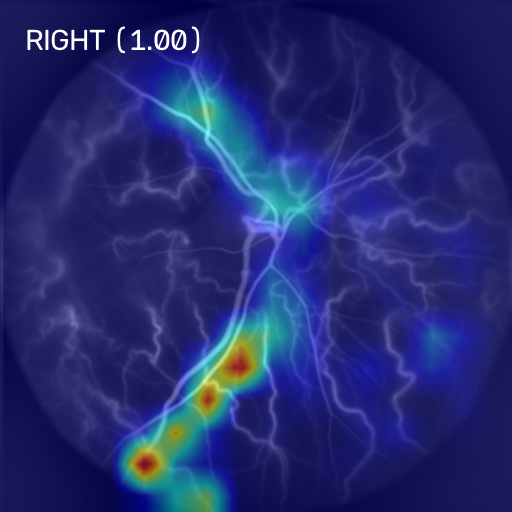}
        \subcaption{Right eye}
    \end{subfigure}
    \hfill
    \begin{subfigure}[b]{0.24\textwidth}
        \centering
        \includegraphics[width=\linewidth]{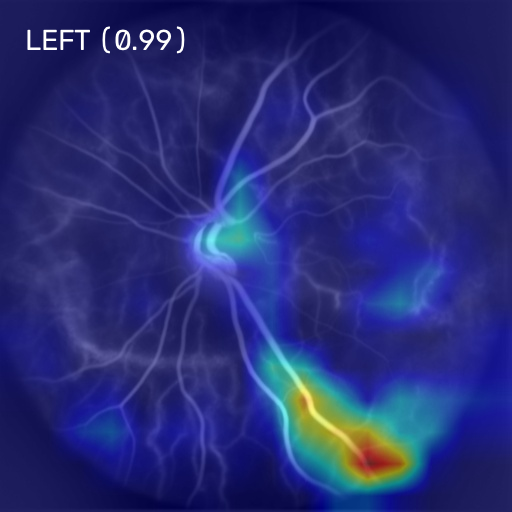}
        \subcaption{Left eye}
    \end{subfigure}
    
    \vspace{0.2cm} 
    
    \begin{subfigure}[b]{0.24\textwidth}
        \centering
        \includegraphics[width=\linewidth]{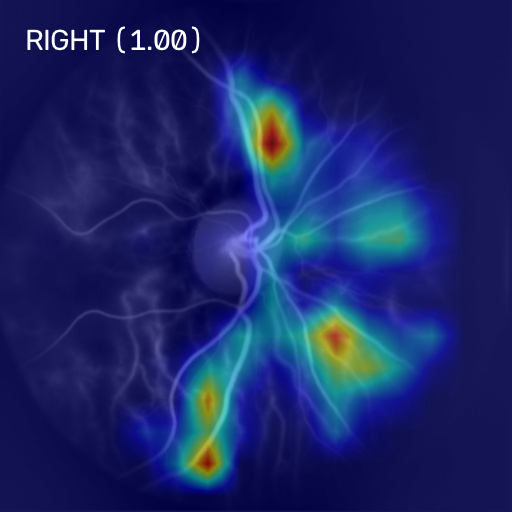}
        \subcaption{Right eye}
    \end{subfigure}
    \hfill
    \begin{subfigure}[b]{0.24\textwidth}
        \centering
        \includegraphics[width=\linewidth]{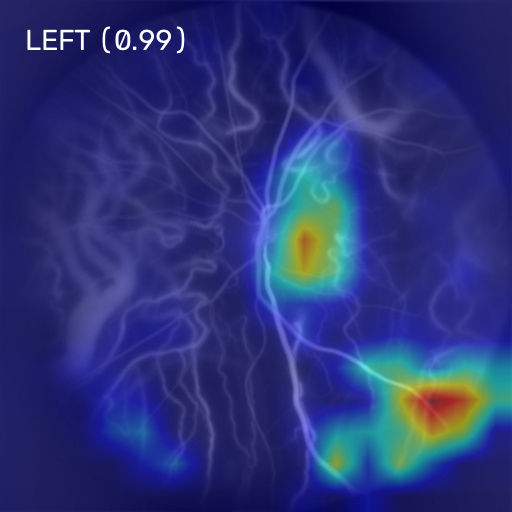}
        \subcaption{Left eye}
    \end{subfigure}
    \hfill
    \begin{subfigure}[b]{0.24\textwidth}
        \centering
        \includegraphics[width=\linewidth]{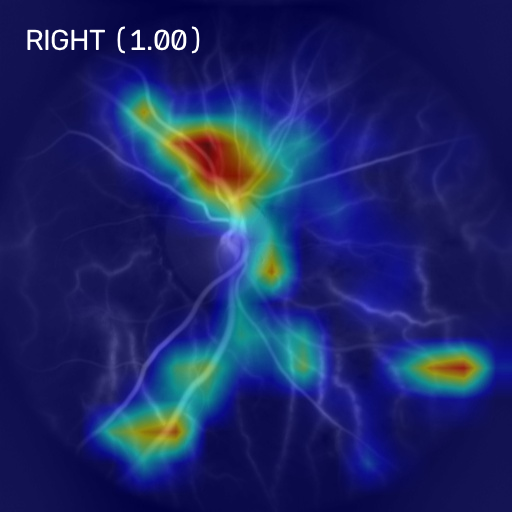}
        \subcaption{Right eye}
    \end{subfigure}
    \hfill
    \begin{subfigure}[b]{0.24\textwidth}
        \centering
        \includegraphics[width=\linewidth]{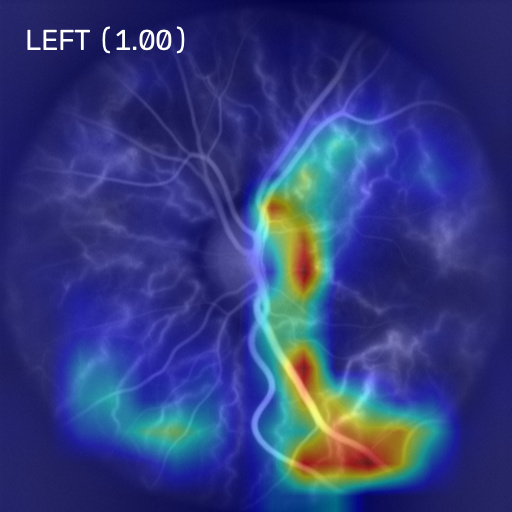}
        \subcaption{Left eye}
    \end{subfigure}
    
    \caption{LayerCAM visualizations of correctly classified M0 images. The activation maps highlight image regions associated with the model's laterality predictions, with warmer colors (red and yellow) indicating stronger activation. The most prominent activations are concentrated around the optic disc and along retinal vessels, suggesting that these anatomical structures contribute to the classification decision.}
    \label{fig:layercam_correct}
\end{figure}

\begin{figure}[!ht]
    \centering
    \captionsetup[sub]{font=footnotesize}
    \begin{subfigure}[b]{0.48\textwidth}
        \centering
        \begin{minipage}{0.49\linewidth}
            \centering
            \includegraphics[width=\linewidth,height=0.18\textheight,keepaspectratio]
                {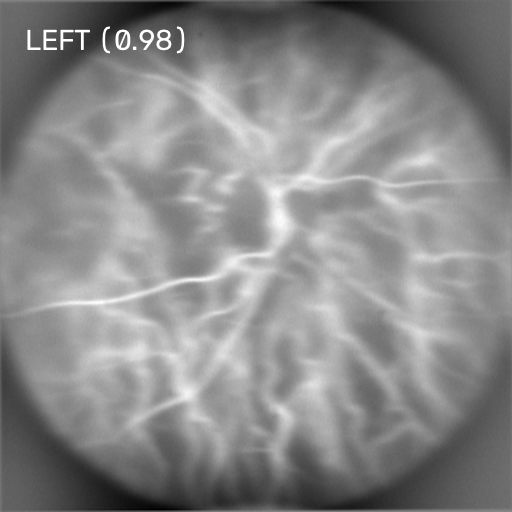}
        \end{minipage}\hfill
        \begin{minipage}{0.49\linewidth}
            \centering
            \includegraphics[width=\linewidth,height=0.18\textheight,keepaspectratio]
                {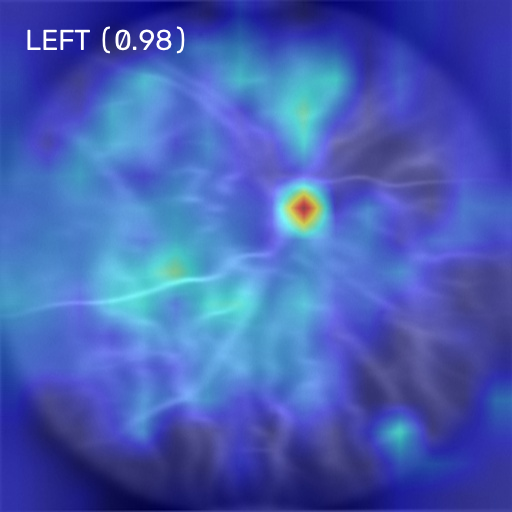}
        \end{minipage}
        \caption{Right eye misclassified; blurred image}
        \label{fig:failure_2}
    \end{subfigure}
    \hfill
    \begin{subfigure}[b]{0.48\textwidth}
        \centering
        \begin{minipage}{0.49\linewidth}
            \centering
            \includegraphics[width=\linewidth,height=0.18\textheight,keepaspectratio]
                {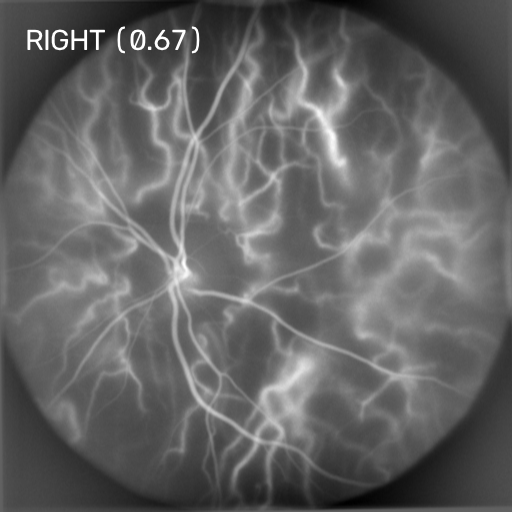}
        \end{minipage}\hfill
        \begin{minipage}{0.49\linewidth}
            \centering
            \includegraphics[width=\linewidth,height=0.18\textheight,keepaspectratio]
                {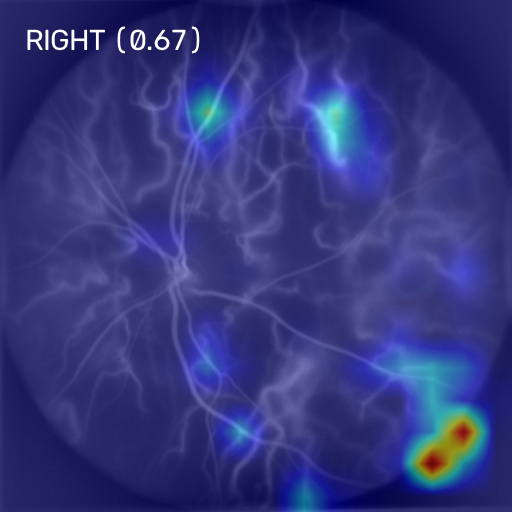}
        \end{minipage}
        \caption{Left eye misclassified; off-center optic disc}
        \label{fig:failure_1}
    \end{subfigure}
    \caption{The two misclassified test images. (a) A severely blurred image, misclassified with high confidence. (b) A sharp image with an off-center optic disc, misclassified with lower confidence. Each M0 image is shown alongside its LayerCAM visualization, with the predicted laterality and confidence score. These maps may help explain the errors but do not establish their cause.}
    \label{fig:m0_failure}
\end{figure}

To interpret the model's predictions, we applied LayerCAM~\cite{jiang2021layercam}, a class activation mapping method that highlights the image regions contributing most to the predicted class. 
As shown in Figure~\ref{fig:layercam_correct}, the activation maps of correctly classified images are mainly concentrated around the optic disc and the retinal vessels, suggesting that the model relies on these anatomical structures to determine laterality. 
In contrast, Figure~\ref{fig:m0_failure} shows the two misclassified images. In the blurred image (Figure~\ref{fig:failure_2}), the strongest activation is a small spot near the optic disc, with diffuse activation across the retina. In the image with the off-center optic disc (Figure~\ref{fig:failure_1}), the strongest activation is near the lower-right border, far from the optic disc. In both cases, the activation does not clearly follow the main retinal vessels.

\section{Discussion}
The high test accuracy of \textit{YOLO11n-cls} indicates that M0 images contain anatomical asymmetries, likely related to vascular geometry and optic disc position, that support laterality classification without explicit fovea detection. This is particularly useful in LDH acquisitions where the fovea lies outside the field of view.

The two misclassifications highlight a limitation of the proposed approach: classification confidence does not always reflect prediction reliability. The blurred image was misclassified with high confidence (0.98), whereas the image with an off-center optic disc received a lower score (0.67). Confidence scores alone may therefore be insufficient to identify unreliable classifications. In the existing LDH workflow, acquisitions that do not provide reliable Doppler measurements are discarded and repeated, which may reduce errors related to poor image quality. However, this would not prevent errors related to an off-center optic disc. Additional evaluation is required to determine whether quality control reliably identifies images at risk of laterality misclassification.

This model is integrated into \textit{DopplerView} version 1.8.1 as an automated preprocessing step. When a new Doppler acquisition is rendered, the system generates the M0 image and immediately predicts the eye side. This prediction is used to:
\begin{enumerate}
    \item Automatically populate the metadata field, reducing the operator's workload.
    \item Serve as a quality control flag if the prediction conflicts with the manual entry.
\end{enumerate}

The dataset was acquired at a single clinical center. Future work will also include eyes with severe anatomical alterations to assess and improve classification robustness.

\section{Conclusion}

Automated eye laterality classification using \textit{YOLO11n-cls} provides a promising solution for retinal laser Doppler holography data management. Achieving 97.62\% accuracy on the held-out test set, even in the absence of visible foveal landmarks, this approach enables automatic assignment of eye laterality metadata while providing an additional mechanism for detecting potential labeling errors. By reducing reliance on manual data entry, the proposed method supports the reliability and scalability of LDH data-processing pipelines.

\section*{Code and data availability}
Model weights (\texttt{.pt}, \texttt{.onnx}) and an example dataset are available on
\href{https://huggingface.co/DigitalHolography/YOLO_EyeSide}{Hugging Face}.
The Python training and visualization code is available on
\href{https://github.com/DigitalHolography/LeftRightEyeClassification}{GitHub}.
Rendering and analysis use \href{https://github.com/DigitalHolography/HoloDoppler}{\textit{HoloDoppler}}.
The classifier is implemented in \href{http://dopplerview.org/}{\textit{DopplerView}} (\href{https://github.com/DigitalHolography/DopplerView/releases/tag/1.8.1}{version 1.8.1}).

\bibliographystyle{unsrt}
\bibliography{references}

\end{document}